\pdfoutput=1

\documentclass[11pt]{article}

\usepackage{acl}

\usepackage{times}
\usepackage{latexsym}

\usepackage[T1]{fontenc}

\usepackage[utf8]{inputenc}

\usepackage{microtype}

\usepackage{inconsolata}

\usepackage{graphicx}

\title{dzNLP at NADI 2024 Shared Task: Multi-Classifier Ensemble with Weighted Voting and TF-IDF Features}

\author{Mohamed Lichouri \\
  LCPTS-FGE. USTHB \\
  Algiers-ALGERIA \\
  \texttt{mlichouri@usthb.dz} \\\And
  Khaled Lounnas \\
  CRSTDLA \\
  Algiers-ALGERIA \\
  \texttt{k.lounnas@crstdla.dz} \\\AND
  Zahaf Boualem Nadjib and Rabiai mehdi ayoub \\
  University of Algiers 01\\
  Algiers-ALGERIA\\
  \texttt{\{zahafndjib,Rabiaimehdiayoub\}@gmail.com}
 \\}

\begin{document}
\maketitle
\begin{abstract} 
This paper presents the contribution of our dzNLP team to the NADI 2024 shared task, specifically in Subtask 1 - Multi-label Country-level Dialect Identification (MLDID) (Closed Track). We explored various configurations to address the challenge: in Experiment 1, we utilized a union of n-gram analyzers (word, character, character with word boundaries) with different n-gram values; in Experiment 2, we combined a weighted union of Term Frequency-Inverse Document Frequency (TF-IDF) features with various weights; and in Experiment 3, we implemented a weighted major voting scheme using three classifiers: Linear Support Vector Classifier (LSVC), Random Forest (RF), and K-Nearest Neighbors (KNN).

Our approach, despite its simplicity and reliance on traditional machine learning techniques, demonstrated competitive performance in terms of F1-score and precision. Notably, we achieved the highest precision score of 63.22\% among the participating teams. However, our overall F1 score was approximately 21\%, significantly impacted by a low recall rate of 12.87\%. This indicates that while our models were highly precise, they struggled to recall a broad range of dialect labels, highlighting a critical area for improvement in handling diverse dialectal variations.
\end{abstract}

\section{Introduction}
\label{intro}
Arabic, with its rich linguistic diversity encompassing numerous dialects alongside Modern Standard Arabic (MSA), presents a significant challenge for natural language processing (NLP) tasks. Despite its importance, many Arabic dialects remain understudied due to resource constraints such as limited research funding and datasets. Addressing this gap, the Nuanced Arabic Dialect Identification (NADI) shared task series \cite{abdul-mageed-etal-2020-nadi,abdul-mageed-etal-2021-nadi,abdul-mageed-etal-2022-nadi,abdul-mageed-etal-2023-nadi} serves as a pivotal initiative to overcome these hurdles. By providing extensive datasets and modeling opportunities, NADI aims to facilitate dialect identification and processing tasks, thereby advancing the understanding and utilization of Arabic dialects.

In the context of feature extraction for Arabic text analysis, standard approaches frequently employ Term Frequency-Inverse Document Frequency (TF-IDF). Our previous work \cite{abbas2019st} in the MADAR'2019 shared task \cite{bouamor-etal-2019-madar} demonstrated the effectiveness of using TF-IDF for efficient feature extraction in Arabic text classification. Drawing inspiration from this prior research, we adopted a similar strategy in our current study. Specifically, in our first experiment, we employed a union of TF-IDF features with various n-gram analyzers (word, character, and character with word boundaries) to enhance the representation of Arabic text.

Moreover, the utilization of weighted feature fusion has emerged as a promising technique for enhancing classification accuracy. \cite{lichouri-etal-2021-arabic} demonstrated the efficacy of weighted feature fusion in multilingual text classification, as showcased in our earlier study in NADI'2023 for Arabic dialect identification \cite{lichouri2023usthb}, emphasizing its adaptability across various linguistic datasets. This methodology closely corresponds to our Experiment 2, which focuses on the weighted concatenation of TF-IDF features.

These earlier studies, along with recent advancements in TF-IDF-based feature extraction and weighted feature fusion, have collectively inspired innovative ideas. In our third experiment, we implemented a weighted hard major voting approach.

This paper, instead of introducing innovative solutions or groundbreaking insights for NADI 2024 \cite{abdul-mageed-etal-2024-nadi}, serves as a concise consolidation of existing knowledge. The rest of the paper is structured as follows: Section \ref{data} provides an overview of the dataset used in our study. In Section \ref{aprch1}, we present our proposed system, followed by discussing the findings and their significance in Section \ref{res}. Finally, our paper concludes in Section \ref{conc}, summarizing the key takeaways and contributions.

\begin{figure*}[!h]
    \centering
    \includegraphics[scale=0.7]{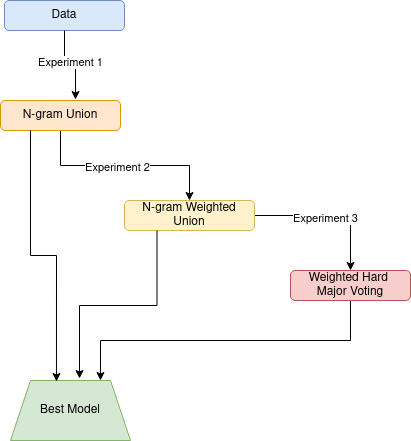}
    \caption{Proposed system for Arabic Dialect Identification.}
    \label{fig1}
\end{figure*}

\section{Description of the Dataset}
\label{data}
The Nuanced Arabic Dialect Identification (NADI) 2024 shared task provided a valuable dataset for Subtask 1: Multi-label Country-Level Dialect Identification (MLDID). The dataset encompassed dialects from various Arabic-speaking countries, including  Egyptian, Saudi Arabian, Algerian, Syrian, Palestinian, and Lebanese.

The dataset provided for MLDID consisted of two parts: a development set, comprising 100 samples, served as a training ground for system development and evaluation, and a test set, containing 1,000 samples, was used for the final evaluation of participant models.

Additionally, participants were granted access to NADI datasets from 2020, 2021, and 2023 for training purposes. These datasets provided a larger pool of Arabic text for system development and refinement.

\section{Proposed system} 
\label{aprch1}
Our proposed system is based on two concatenation or union process of features vs classifiers where we opted for three experiments:
\begin{itemize}
    \item \textbf{Experiment 1 \cite{lichouri2018word, abbas2019st}:} In this experiment, we employ the TF-IDFVectorizer, which employs three analyzers (Word, Char, and Char\_wb), each with varying n-gram ranges. In the default configuration, we combine these three features, assigning equal weights of 1 to each. During feature extraction, we varied the n-gram values (ranging from $n=1$ to $5$). Finally, the SVC classifier was trained.
    \item \textbf{Experiment 2 \cite{lichouri-etal-2021-arabic, lichouri2023usthb}:} In this instance, we combine the three TF-IDF features using a weight vector comprising three distinct values (w1, w2, w3) corresponding to the Word, Char, and Char\_wb analyzer, respectively. The LSVC classifier was then trained.
    \item \textbf{Experiment 3:} In this instance, we applied a weighted hard major voting for three classifier (LSVC, RF and KNN).
\end{itemize}

\begin{table}[!h]
\caption{The various parameters used in our work}
\label{params}
\begin{tabular}{|c|c|}
\hline
\textbf{Settings}   & \textbf{Values}        \\ \hline
N-GRAM       & (1,n) with n=1 to 5 \\ \hline
TFIDF      & \begin{tabular}[c]{@{}l@{}} transformer\_weights: 0.1 - 1 \\ max\_features: 300-1000\end{tabular}             \\ \hline
LSVC                 & \begin{tabular}[c]{@{}l@{}}C= 1 - 5 and\\ class\_weight='balanced'\end{tabular} \\ \hline
RF & Default\\ \hline
KNN & n\_neighbors=3 \\ \hline
Major Voting & weight = 0.1 - 0.6\\\hline 

\end{tabular}
\end{table}

\begin{table*}[!h]
\centering
\caption{Obtained F1-score from merging TF-IDF representations with uniform and different weighting}
\label{tab2}
\begin{tabular}{|c|l|c|c|c|c|c|}
\hline
Id       & n-gram range                                                  & Classifier Configuration           & Transformer weights                                                                                            & F1-score \\ \hline
1 & (3, 5, 5) & class\_weight='balanced', C=5 &                                                                                                  & 20.64   \\ \hline
2 & (5, 5, 5) & class\_weight='balanced', C=4 & tw=\{0.65, 0.85, 0.85\}  & 22.51   \\ \hline
3 & (3, 4, 5)& C=4                                & tw=\{0.45, 0.5, 0.75\}   & 21.07   \\ \hline
4  & (4, 4, 4) & C=4                                & tw=\{0.45, 0.5, 0.75\}   & 20.78   \\ \hline
5  & (4, 4, 4) & C=4                                & tw=\{0.35, 0.45, 0.75\} & 20.51   \\ \hline
\end{tabular}
\end{table*}

The various parameters used in these three experiments are reported in Table \ref{params} where we specify the settings and corresponding values for each parameter. For N-GRAM, we explored n-gram ranges from unigrams to 5-grams. In TFIDF, we varied the transformer weights from 0.1 to 1 and the maximum number of features from 300 to 1000. Additionally, for the LSVC classifier, we adjusted the regularization parameter C from 1 to 5 and set the class weight to 'balanced'. RF classifier was used with default settings, while for KNN, we set the number of neighbors to 3. Lastly, in the Major Voting ensemble technique, we experimented with weights ranging from 0.1 to 0.6.

\section{Obtained results}
\label{res}
During our participation in the MLDID (Closed Track) project, we iteratively explored various approaches to enhance our multi-label dialect identification models (see Table \ref{fig1}). This exploration encompassed feature engineering with TF-IDF, incorporating ensemble methods, and experimenting with different voting combination strategies.

As a baseline experiment, we utilized a basic TF-IDF representation with 1-gram features and a linear kernel SVC classifier. Despite its simplicity, it achieved a low F1-score of 19.43\%. 

In the first experiment, we expanded the feature space by incorporating a combination of word and character-level n-grams (word, char and char\_wb) in TF-IDF representation. We employed the Linear Support Vector Classifier (LSVC) classifier with balanced class weights, resulting in notable improvements in F1-score of 20.64\%. 

In the second experiment, we experimented with a different combination of n-grams while maintaining the balanced class weights. We also explored variations in the weights assigned to different feature types within the TF-IDF representation. The obtained F1-score ranged from 20.51\% to 22.51\% (see Table \ref{tab2}).

In the third experiment, we introduced ensemble methods, namely hard voting and weighted hard voting, combining SVC, RF, and KNN classifiers. Interestingly, despite the complexity of these ensemble approaches, their F1-score were lower compared to well-tuned single LSVC configurations (16.33\% and 21.44\%, respectively). This suggests that the ensemble methods may not have effectively leveraged the complementary strengths of the individual classifiers or their voting strategies might not have been optimal for this task.

\section{Conclusion} 
\label{conc}
Our analysis of the NADI 2024 shared task highlights the critical role of feature engineering and model optimization in Arabic dialect identification. Key findings include integrating character-level features alongside word-level features consistently improved performance, balanced class weights within the LSVC classifier significantly enhanced F1-score, surpassing 20.64\%, strategic assignment of transformer weights yielded the highest F1-score of 22.51\%, and ensemble methods like hard voting achieved moderate F1-score but were surpassed by finely tuned single SVC configurations. In conclusion, our study emphasizes the importance of incorporating character-level information, utilizing balanced class weights, and exploring advanced feature weighting techniques to advance Arabic dialect identification systems. These insights provide valuable guidance for future research in this domain.

\bibliography{custom}
\bibliographystyle{acl_natbib}

\end{document}